\documentclass[11pt]{article} 
\usepackage{times}

\usepackage[margin=1in]{geometry}

\usepackage{microtype}
\usepackage{graphicx}
\usepackage{subfigure}
\usepackage{booktabs} 

\usepackage{hyperref}

\usepackage{amsmath}
\usepackage{amssymb}
\usepackage{mathtools}
\usepackage{amsthm}
\usepackage{algorithm}
\usepackage{algpseudocode}
\usepackage{natbib}
\usepackage{listings}

\theoremstyle{plain}

\theoremstyle{definition}

\theoremstyle{remark}

\usepackage[textsize=tiny]{todonotes}

\usepackage{tabularx}
\usepackage{booktabs}
\usepackage{enumitem}
\usepackage{soul}
\usepackage{tcolorbox}
\tcbuselibrary{breakable}

\usepackage{adjustbox}

\newcommand{\np}{\mathsf{NP}}
\newcommand{\conp}{\mathsf{co{\text -}NP}}
\newcommand{\p}{\mathsf{P}}
\newcommand{\pspace}{\mathsf{PSPACE}}
\newcommand{\expt}{\mathsf{EXPTIME}}

\renewcommand{\paragraph}[1]{\vspace{.05in}\noindent\textbf{{#1}.~~}}
\usepackage[noabbrev,capitalize,nameinlink]{cleveref}
\usepackage{enumitem}
\usepackage{graphicx}
\usepackage{multirow}
\usepackage{siunitx}
\usepackage{subcaption}
\usepackage{wrapfig}
\usepackage{xcolor}

\definecolor{blush}{rgb}{0.87, 0.36, 0.51}

\title{PEA: Enhancing LLM Performance on Computational-Reasoning Tasks}

\author{\small 
  \hspace{-2em} Zi Wang$^1$\footnote{Equal Contribution} \vspace{-0.4em}\\ \hspace{-2em}\scriptsize zw@cs.wisc.edu
  \and \hspace{-2em} \small Shiwei Weng$^{2*}$\vspace{-0.4em}\\\hspace{-2em}  \scriptsize weng@cs.jhu.edu
  \and \hspace{-2em} \small Mohannad Alhanahnah$^3$ \vspace{-0.4em}\\ \hspace{-2em}\scriptsize mohannad.alhanahnah@chalmers.se
  \and \hspace{-2em} \small Somesh Jha$^1$ \vspace{-0.4em}\\ \hspace{-2em}\scriptsize jha@cs.wisc.edu
  \and \hspace{-2em} \small Tom Reps$^1$ \vspace{-0.4em}\\ \hspace{-2em}\scriptsize reps@cs.wisc.edu\\
}
\date{\footnotesize 1. Department of Computer Sciences, University of Wisconsin-Madison\\
 \footnotesize 2. Department of Computer Science, Johns Hopkins University\\
\footnotesize 3. Department of Computer Science and Engineering, Chalmers/University of Gothenburg \\}

\begin{document}
\maketitle
\begin{abstract}
Large Language Models (LLMs) have exhibited remarkable capabilities across diverse domains, prompting investigations into their potential as generic reasoning engines. While recent studies have explored inference-time computation to enhance model performance on complex problems, current research lacks a formal framework to characterize the complexity of reasoning tasks. This study introduces the Predicate-Enumeration-Aggregation (PEA) framework, a formal approach to describe and solve a class of important reasoning tasks termed \emph{computational reasoning} problems. The PEA framework decomposes these problems into predicate and enumeration components, using LLMs to synthesize programs based on specified predicates, enumeration, and aggregation rules. These synthesized programs are then executed to obtain solutions to the computational tasks. We demonstrate the framework's efficacy on benchmark tasks including Boolean satisfiability problems, game of $24$, and planning problems. Empirical evaluation reveals that PEA substantially enhances the performance of underlying models on benchmark computational problems, yielding an average accuracy improvement of approximately $50\%$, coupled with increased efficiency.
\end{abstract}
\section{Introduction}

Large Language Models (LLMs) have exhibited significant generalization capabilities across diverse domains, prompting investigations into their potential as generic reasoning engines. Recent studies have explored inference-time computation techniques~\citep{welleck2024from,snell2024scalingllmtesttimecompute}, particularly prompt engineering methods such as Chain-of-Thought (CoT), to enhance LLM performance on complex reasoning tasks~\citep{wei2022chain}. These approaches have successfully improved model performance and expanded LLMs' practical applications. However, despite the growing focus on enhancing model capabilities through inference-time computation for complex reasoning tasks, the current literature lacks a formal framework to precisely describe and characterize the complexity of reasoning problems.

This study identifies a class of reasoning problems, termed \emph{computational reasoning} problems, which are particularly challenging for LLMs~\citep{yao2023tree,hao2024planningrigorgeneralpurposezeroshot,valmeekam2023planbench},
such as planning problems and arithmetic games.
Informally, these problems can be accurately described using succinct programmatic representations.
We propose a formal framework to describe and algorithmically solve these problems.

The framework employs first-order logic, equipped with efficiently computable predicates and finite domains. A first-order logic formula~\citep{kleene1971introduction},
is represented syntactically as:
\[
Q_1 x_1 Q_2 x_2\ldots Q_n x_n \ldotp P(x_1, \ldots, x_n),
\]
where $Q_i$'s are quantifiers and $P(\cdot)$ is a predicate.

To illustrate, the statement ``The sum of 
an odd number and an even number
is odd'' can be expressed as:
\begin{equation*}
\begin{split}
(\forall x, y, z ) (\exists l, m, n ) \ldotp & [x = 2l \land y = 2m+1 \land z = x+y] \\ &\rightarrow z=2n+1
\end{split}
\end{equation*}
where all $x, y, z, l, m, n \in \mathbb{Z}$, the integer set.

As used in this paper, additional
restrictions are enforced, limiting variables to finite domains and ensuring predicates are efficiently evaluable. While these limitations preclude the representation of certain mathematical propositions, such as those involving infinite sets like $\mathbb{Z}$, they facilitate algorithmic resolution of problem correctness through exhaustive enumeration and aggregation of solution candidates. Despite reduced expressiveness, this approach remains sufficiently powerful for describing numerous real-world problems (see~\cref{sec:comp}). To illustrate, we consider the Boolean Satisfiability (SAT) problem~\citep{Karp1972}, formally expressed as:
\[
\exists x\in\{0, 1\}^n\ldotp B(x)=\text{True},
\]
where $B(x)$ represents a Boolean logic formula.  The SAT problem determines whether a satisfying assignment exists for $B$.
\begin{figure*}
    \centering
    \includegraphics[width=0.75\textwidth]{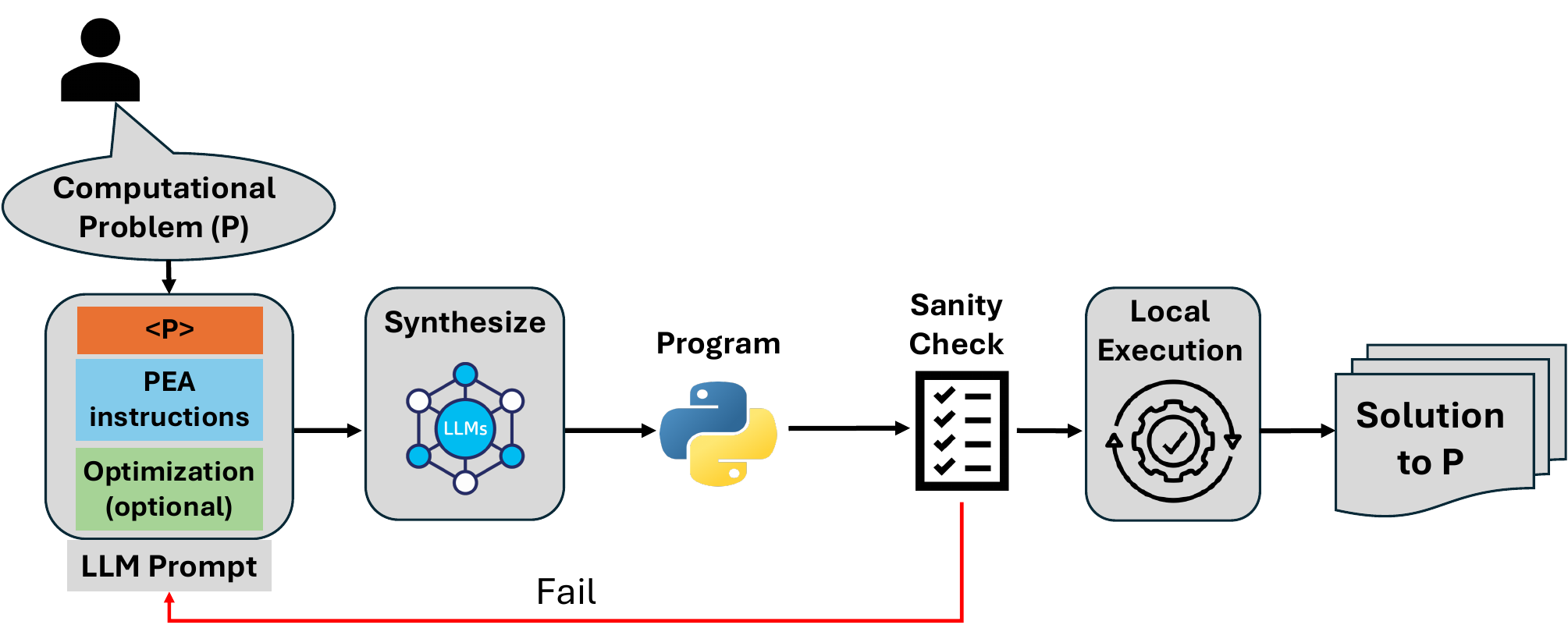}
    \caption{PEA Pipeline: Given a problem $P$ described in natural language, the PEA section of the prompt instructs the LLM to synthesize problem-solving programs. Programs passing sanity checks are executed locally to obtain solutions for $P$. Users have the option to query the LLM with $P$ to obtain search-space optimization strategies.}\label{fig:pipeline}
    \vspace{-7pt}
\end{figure*}
To ascertain the validity of $\exists x\in\{0, 1\}^n\ldotp B(x)=\text{True}$, a brute-force approach enumerates all possible candidates $c$ from $\{0, 1\}^n$ and evaluates $B(c)$.

Given this formal framework for describing computational problems, we propose a method to solve real-world problems using LLMs (see~\cref{sec:algo}). 
Our approach employs
an algorithm that performs
exhaustive enumeration of
valid candidate values for variables bounded by quantifiers,
and comprehensive predicate evaluation. This strategy inherently aligns with programmatic implementation, leveraging programs' intrinsic capacity for precise evaluation and thorough enumeration when correctly formulated.
We introduce the Predicate-Enumeration-Aggregation (PEA) framework, which operates as follows:
(i) the predicate, enumeration,
and aggregation rules are described in natural language;
(ii) LLMs are instructed to generate programs that fulfill these natural-language tasks; and
(iii) the synthesized programs are executed to determine the validity of the quantified predicate (see~\cref{sec:prog,sec:implementation}).
\Cref{fig:pipeline} illustrates the pipeline of the PEA framework.

This algorithmic framework can be conceptualized as employing a brute-force, trial-and-error approach to problem-solving. Due to the expressiveness of PEA, the algorithm may be optimal asymptotically given the hypothesized essential lower-bounds in many computational problems~\citep{SETH,fgc}, potentially necessitating extensive enumerations. The PEA framework leverages computational efficiency in programmatic predicate evaluation, facilitating rapid assessment of millions of simple predicates within seconds. This approach capitalizes on the high performance of modern computer architectures and code optimization techniques, rendering execution computationally inexpensive and expanding the feasible search space. Thus, this methodology is applicable to various real-world applications with moderate search spaces. In this context, programs function as reasoning generators, with their execution producing concrete reasoning outcomes.
One motivation is that for large problem instances, the actual reasoning often exceeds the algorithmic description in length, due to the hypothetical computational lower bound. This phenomenon facilitates easier synthesis of reasoning generators compared to direct reasoning. Furthermore, the framework's efficiency can be enhanced through strategic pruning of ineffective candidates within the enumeration space. 
Consequently, the PEA framework transforms lengthy enumerations into succinct programmatic representations, augmenting LLMs' reasoning capabilities by leveraging their coding proficiency. 

In summary, our work makes the following contributions: 
\begin{itemize}[parsep=0pt,itemsep=1pt,leftmargin=13pt]
    \item The development of PEA, a formal framework grounded in first-order logic, which facilitates the categorization of computational-reasoning problems.
    \item The introduction of a novel prompt-engineering technique, derived from the PEA framework, designed to augment the capabilities of LLMs in addressing complex computational tasks via their coding capability.
    \item  Empirical evaluation demonstrating that PEA substantially enhances the performance of underlying models, yielding an average accuracy improvement of approximately $50\%$ while enhancing efficiency.
\end{itemize}
\section{Related Work}
First-order logic, initially championed by Hilbert as the formal language for mathematical description, formed the basis of his ambitious formalism project in mathematics~\citep{Hilbert1926}. This endeavor was subsequently challenged by Gödel and Turing through their seminal work on first-order arithmetic and computability theory~\citep{Goedel1931,turing1936computable}.~\citet{Kleene-recursive} further extended Turing's halting problem construction to develop the arithmetical hierarchy using alternating quantified computable predicates. Computational complexity theorists later adapted ideas from Turing, Gödel, and Kleene to establish foundational results in time and space complexity, catalyzing systematic studies in complexity theory and algorithm design~\citep{arora_complexity,algo}. Our proposed formalism can be viewed as an adaptation of these computational constructions to model real-world reasoning problems. This approach enables formal categorization of reasoning problems and potentially introduces additional theoretical tools for their analysis.

LLMs have demonstrated significant potential across various domains, particularly through the application of test-time computation to enhance their performance. Prompt engineering techniques, such as CoT and Tree of Thoughts (ToT)~\citep{wei2022chain,yao2023tree}, have been shown to augment the reasoning capabilities of less robust LLMs on complex reasoning tasks. However, a formal framework for
classifying
the complexity of reasoning problems has been lacking. This study proposes a formal description of reasoning, which aims to facilitate researchers' understanding of the underlying complexity of reasoning problems.

We employ LLMs to synthesize programs that implement our PEA framework, addressing real-world computational problems. While recent works have utilized LLMs for program synthesis in reasoning tasks, our approach distinguishes itself by targeting more complex reasoning problems~\citep{pal,weir2024learning}. From a first-order-logic perspective, we 
go beyond quantifier-free formulations used in
previous studies to quantified problems (see appendix for a detailed discussion).

A parallel line of research uses established solvers as underlying reasoning engines for real-world problems, with LLMs facilitating the translation process~\citep{liu2023llmp,hao2024planningrigorgeneralpurposezeroshot}. However, a programming language constrained by solver binding significantly reduces its programmability and often necessitates substantial human effort to ensure translation correctness, particularly in API usage and adherence to specific programming structures. Furthermore, while solvers typically use fixed proof-search optimization strategies, our approach leverages LLMs' programmability to address a broader range of challenges and allows customization of search space optimization, potentially offering greater flexibility and efficiency in problem-solving. 
\section{Method}
\label{Se:Method}

This section provides a detailed exposition of the PEA framework, encompassing four key aspects: syntax (formal structure), semantics (real-world problem modeling), computation (algorithmic problem-solving approaches), and implementation (practical realization). 

\subsection{Notation and definitions}\label{sec:def}
Let $X$ be a finite set. Predominantly, we consider $X=\{0,1\}$, the binary set. We use $x$ to denote variables, and $c$ to denote concrete values. Central to our analysis is the concept of an $n$-ary \emph{unquantified} predicate, denoted as $P: X^n\rightarrow \{0, 1\}$. This predicate function serves to determine whether $n$ variables satisfy the relation specified by $P$. For instance, $P(x, y): x+y = 5$ denotes
the set of $(x,y)$ pairs such that the sum of the values of $x$ and $y$ equals $5$---e.g.,
$P(1, 4)=1$ and $P(1,2) = 0$. We introduce the existential ($\exists$) and universal ($\forall$) quantifiers, denoted by the set $Q=\{\exists, \forall\}$. The logical connectives $\land$, $\lor$ and $\neg$ represent ``\textbf{and}'', ``\textbf{or}'' and ``\textbf{not}'' operations, respectively. Quantifiers 
bind
variables within predicates, forming quantified predicates of the form:
\begin{equation}\label{eq:qp}
    Q_1 x_1 Q_2 x_2\ldots Q_n x_n \ldotp p(x_1, \ldots, x_n),
\end{equation}
where $Q_i\in Q$. Variables bound by quantifiers are termed \emph{bounded}; while those not bound are \emph{free}. A predicate with all variables substituted by concrete values is called a \emph{concrete} predicate.

The validity of quantified predicates serves as an abstraction for validating real-world problems. In the case of predicates where all variables are substituted with specific values, the validity can be unambiguously determined through evaluation of the concrete predicate. However, predicates containing free variables have indeterminate validity. For example, $x + 2 = 5$ is neither true nor false due to the unknown value of $x$. The validity of fully bounded quantified predicates can be ascertained within a predefined domain. For instance, given $X = \{2,3\}$, $\exists x\ldotp x+2=5$ is true while $\forall x\ldotp x+2=5$ is false. Conversely, if $X=\{3\}$, then $\forall x\ldotp x+2=5$ is true.

\subsection{Computational problems}\label{sec:comp}
In this section, we introduce an abstraction of computational problems through the framework of quantified predicates. We focus our attention on~\Cref{eq:qp}, where $P(x_1, \ldots, x_n)$ is an efficiently decidable predicate. The notion of efficient decidability in this context implies that for any substituted values $c_1, \ldots, c_n$, the evaluation of $P(c_1, \ldots, c_n)$ can be executed within polynomial time relative to the encoding size of $c_1, \ldots, c_n$.
As is well known,
the formulation presented in \Cref{eq:qp} possesses sufficient expressive power to abstract several well-known classical computational problems
\cite{DBLP:books/daglib/0095988}.

\paragraph{$\np$ class}The $\np$ complexity class represents a fundamental category in computational-complexity theory. Problems within $\np$ can be formulated as decision problems concerning the validity of existentially quantified predicates. A paradigmatic example of this class is SAT, which holds a central position as an $\np$-complete problem. The SAT problem can be expressed in the following form:
\begin{equation}\label{eq:np}
   \exists x\ldotp P(x), 
\end{equation}

where $P(x): \phi(x)=1$ denotes a Boolean formula $\phi$ evaluating to true. In this formulation, the SAT problem encapsulates the question of whether, for any given Boolean formula $\phi$, there exists a solution that renders $\phi$ true.

\paragraph{$\conp$ class}A complexity class closely related to $\np$ is the $\conp$ class. While $\np$ problems are characterized by existential quantification, $\conp$ problems are formulated using universal quantification, represented as:
\[
\forall x\ldotp P(x),
\]
where $P(x): \phi(x)=1$ denotes an arbitrary Boolean formula $\phi$ evaluating to true. The canonical $\conp$ problem is the \emph{tautology} problem, which inquires whether a given Boolean formula holds true for all possible assignments. 

In fact, more complex quantifier patterns beyond single existential or universal quantifiers can be employed to bound variables. These predicates lead to computational problems hypothetically harder than $\np$ and $\conp$ problems, such as those in the polynomial hierarchy and $\pspace$-hard problems. For a more comprehensive treatment of these topics, readers are directed to~\citet{arora_complexity}.

\subsection{Algorithmic solution}\label{sec:algo}
Having established~\cref{eq:qp} as a general formulation for computational problems, we introduce an algorithmic approach to determine their validity. Without loss of generality,
Let $X = \{c_1, c_2\}$ and $P: X^2\rightarrow \{0, 1\}$ and assume that the evaluation of each concrete predicate $P(c_i,c_j)$ is computationally efficient. Under these assumptions, it becomes theoretically feasible to exhaustively enumerate all legitimate substitutions to evaluate the validity of the problem. The algorithmic procedure for this approach is outlined as follows:
\begin{enumerate}
[parsep=0pt,itemsep=1pt,leftmargin=13pt]
\item For an existentially quantified predicate:
$\exists x \in X^2 \ldotp P(x) = P(c_1, c_1) \lor P(c_1, c_2) \lor P(c_2, c_1) \lor P(c_2, c_2)$.
\item For a universally quantified predicate:
$\forall x \ldotp P(x) = P(c_1, c_1) \land P(c_1, c_2) \land P(c_2, c_1) \land P(c_2, c_2)$.
\end{enumerate}
For more general finite domains and quantifier domains, we can unravel the quantifiers sequentially according to the leading quantifier and ultimately resolve the validity of the quantified predicate based on all concrete predicate evaluations. In complexity theory, this result is known as $\pspace\subseteq\expt$~\citep{arora_complexity}. 

This algorithmic solution represents a worst-case scenario, exhibiting exponential-time complexity. Given the expressive power of predicates in general computational problems,
this approach may be no worse asymptotically than
the theoretical optimum achievable.
The approach thus
aligns with the Exponential Time Hypothesis~\citep{SETH}, which posits that even for the SAT problem—a seemingly simple instance of~\cref{eq:qp}—algorithms cannot surpass the performance of brute-force methods, inevitably resulting in exponential time complexity. 
However, it is noteworthy that in many practical applications, the solution enumeration space is typically constrained to a magnitude of a few hundred thousand, rendering the approach computationally feasible within these bounds. Moreover, in numerous applications, the enumeration space can be optimized through the pruning of ineffective candidates, thereby enhancing the efficiency of the algorithmic process.



\subsection{Program synthesis}\label{sec:prog}
Our objective is to accurately solve real-world computational problems. In accordance with the algorithmic resolution presented in~\cref{sec:algo}, we aim to enumerate all concrete predicates. This approach necessitates the fulfillment of two critical conditions:
\begin{enumerate}
[parsep=0pt,itemsep=1pt,leftmargin=13pt]
\vspace{-0.2cm}
    \item The evaluation of the predicate with specific value assignments must consistently yield accurate results.
    \item The enumeration process must be exhaustive, encompassing all legitimate solution candidates without omission.
\end{enumerate}
To optimize efficiency, additional considerations are warranted. Ideally, the enumeration process should minimize evaluations of repetitive or ineffective candidates. Furthermore, each individual predicate evaluation should be executed in a time-efficient and space-efficient manner.
These requirements naturally suggest
the use of program synthesis.

Contemporary computer architectures offer highly efficient program execution, facilitating precise and complete enumeration with accurate predicate evaluation. Our task thus reduces to synthesizing programs capable of evaluating all enumerations and deriving the validity of quantified predicates as expressed in~\cref{eq:qp}. 
%
Our methodology for Python program synthesis, aligned with our formal framework, comprises three key functions:
\begin{enumerate}
[parsep=0pt,itemsep=1pt,leftmargin=13pt]
\vspace{-0.2cm}
    \item Predicate evaluation: This function assesses whether a concrete solution candidate successfully resolves the problem;
    \item Enumeration synthesis: This function generates all legitimate solution candidates, with the capability to prune those known to be ineffective;
    \item Aggregation of results: This function orchestrates the overall synthesis process. It invokes the enumeration function to generate candidates, utilizes the predicate function to evaluate these candidates, and returns results based on the quantifiers.
\end{enumerate}

We term this approach the Predicate-Enumeration-Aggregation (\textbf{PEA}) framework. LLMs are instructed to synthesize these components, and the resulting programs are executed on specific problem instances to derive solutions.

\begin{algorithm}[tb]
   \caption{PEA Program Synthesis Algorithm}
   \label{alg:pea}
\begin{algorithmic}[1]
   \State {\bfseries Input:} A computational problem $P$, an LLM $M$, an iteration upper bound $m$, and a Boolean option for enumeration optimization $op$.
   
   {\bfseries Output:} Synthesized code $C$ to solve $P$.
   \State Prepare template $T$ incorporating $P$'s description and PEA components.
   \If{$op$ is True}
   \For{$i=1$ {\bfseries to} $m$}
   \State Query $M$ with $P$ for enumeration-space reduction strategy and integrate strategy into $T$ to create augmented template $T'$.
   
   \State Query $M$ to synthesize code $C'$ with $T'$.
   \If{$C'$ passes code integrity check}
    \State \textbf{\emph{Return}} $C'$.
   \EndIf
   \EndFor
   \EndIf
   \For{$i=1$ {\bfseries to} $m$}
   \State Query $M$ to synthesize code $C$ with $T$
   \If{$C$ passes code integrity check}
    \State \textbf{\emph{Return}} $C$.
    \EndIf
    \EndFor
   \State \textbf{\emph{Return}} an empty program.
\end{algorithmic}
\end{algorithm}

\subsection{Implementation}\label{sec:implementation}
The PEA framework implementation employs a structured-prompt methodology, combining natural-language problem descriptions with specific Python code-generation instructions. These instructions detail function names and input/output descriptions. For problem instances that cannot be represented by primitive Python types, the LLM is directed to synthesize compound types, encapsulated as Python classes, to describe the problem structure. 

Users have the option to prompt the model for autonomous generation of pruning instructions during enumeration synthesis, based on the natural language problem description. This approach enhances system efficiency and autonomy in problem-solving tasks, reducing human intervention.

Code integrity is checked through syntactic and semantic validation. Syntactic checks employ Python's abstract syntax tree (AST) module to confirm function presence, input/output types, non-empty definitions, and correct problem-instance conversion. Semantic checking executes the generated code with a fixed example (either provided by or sampled from the dataset) and compares the output to the expected result. Successful code synthesis requires passing all these validation checks, as outlined in~\cref{alg:pea}. It is noteworthy that the synthesis of optimized enumeration functions presents increased complexity due to the intricate nature of pruning instructions. Consequently, in instances where the model fails to successfully synthesize code incorporating these pruning instructions, it defaults to the generation of naive enumeration functions. 

\section{Evaluation}
This section presents empirical evaluations comparing our methodology with existing prompt-engineering techniques, addressing three primary research questions:
\begin{tcolorbox}
\begin{enumerate}[start=1,label={\bfseries RQ\arabic*:}]
\item What is the capacity of LLMs to generate exhaustive enumerations?
\item Could PEA enhance LLMs' reasoning capabilities through program synthesis and local executions?
\item 
Compared to direct reasoning, does PEA achieve more efficient reasoning by combining reasoning with program execution?
\end{enumerate}
\end{tcolorbox}
\textbf{RQ1} examines LLM reasoning limitations for problems with succinct programmatic but extensive end-to-end representations. This inquiry is foundational, because many computational-reasoning problems are hypothesized to have essential computational lower-bounds~\citep{SETH,fgc}, potentially necessitating lengthy enumerations.
\textbf{RQ2} investigates whether the PEA
program-synthesis framework enhances an LLM's reasoning capability on computational tasks by leveraging their coding proficiency,
aiming to determine if programmatic problem translation improves
an LLM's
performance on complex reasoning tasks.
\textbf{RQ3} evaluates the efficiency of the PEA framework, including PEA representation translation and local server execution, to assess whether it offers a more efficient alternative to direct reasoning by LLMs.

\paragraph{Models and server}We use recent language models with advanced coding and reasoning capabilities, specifically GPT-4o, OpenAI o1-mini, and o1 models~\citep{openai2024gpt4technicalreport}. The majority of computational cost 
stems from LLM queries. Local Python execution is performed on a server with thirty-two AMD EPYC 7313P 16-core processors, 528 GB of memory. Supplementary small-scale experiments employing recent DeepSeek models were conducted, with results presented in the appendix~\citep{deepseekai2024deepseekv3technicalreport,deepseekai2025deepseekr1incentivizingreasoningcapability}.

\paragraph{Method and baselines} We implemented \textbf{PEA} as described in~\cref{sec:implementation}, using a fixed example for code integrity checks, which is removed from the evaluation dataset if present. Comparative analysis employed Direct Query (\textbf{DQ}) and \textbf{CoT} as principal baselines, with \textbf{ToT} as an additional baseline for the Game of $24$. Our core PEA prompts are provided in the appendix.

\paragraph{Experimental design}
To address \textbf{RQ1}, LLMs are evaluated on their ability to generate enumerations for combinatorial problems, specifically Cartesian-product combinations and permutations (see \Cref{sec:enum-llm}).
%
The performance metric is the percentage of total possible enumerations successfully outputted by the models.
For \textbf{RQ2}, we selected a range of computational-reasoning tasks, including SAT, Game of $24$, and two planning tasks derived from recent LLM literature~\citep{marino2024fastanalysisopenaio1preview,yao2023tree,valmeekam2023planbench}. PEA and 
various
baselines were applied to find correct answers (see~\cref{sec:sat,sec:gon,sec:planning}). Among all computational tasks, optimization of enumeration in PEA code synthesis is only activated for the logistics planning problem. For~\textbf{RQ3}, the time used by PEA and other tools to solve the entire dataset was measured (see~\cref{sec:eff}). Notably, for each task, PEA's program synthesis is performed only once, with the resulting program being reused across the entire dataset, as each data point represents a concrete problem instance serving as input to the synthesized program.

The PEA-generated code undergoes integrity checks with a maximum of $m=10$ iterations, as 
per~\cref{alg:pea}.
A $30$-second execution timeout is imposed on synthesized code. Solutions are considered unfound if their execution exceeds 
the timeout threshold.

\paragraph{Findings}
Our
experiments yielded
the following results pertaining to the research questions:
\begin{enumerate}[parsep=0pt,itemsep=1pt,leftmargin=28pt,start=1,label={\bfseries RQ\arabic*:}]
\item LLMs exhibit accurate enumeration for small search spaces (typically less than a few hundred elements) but default to generating Python code for larger enumerations rather than providing exhaustive results. This preference for programmatic approaches over end-to-end representations aligns with our premise of leveraging LLMs' coding capabilities for complex reasoning tasks, where code serves as a reasoning generator and its execution produces concrete reasoning outcomes.

\item Our evaluation of computational problems demonstrates that reasoning via code generation and execution consistently outperforms direct reasoning when using identical underlying models. Programs passing integrity checks successfully resolve a substantial portion of the dataset, underscoring the efficacy of programmatic approaches in computational problem-solving. The o1 and o1-mini models significantly surpass GPT-4o in direct reasoning. Moreover, these models exhibit superior proficiency in synthesizing complex code, particularly in generating optimized solutions that adhere to enumeration pruning instructions. This enhanced capability further augments PEA's performance on more intricate tasks.

\item Code synthesis and program execution provide efficient reasoning, capitalizing on decades of optimization in modern computer architectures. This approach offers a significant advantage over recent powerful reasoning models, which, despite their accuracy, require substantially more time for problem-solving. Notably, successful code generation typically necessitates fewer than three LLM queries to pass integrity checks. This minimal query requirement results in negligible computational cost when amortized across the entire dataset, further enhancing the method's overall efficiency.
\end{enumerate}

The following subsections offer a comprehensive analysis of each experimental
task.

\subsection{Enumerations with LLM}\label{sec:enum-llm}
In our evaluation, we employed random 3-character strings as variable-name candidates. For \textbf{Cartesian product-combination} tasks, LLMs were instructed to output $n$-variable product expressions, producing all possible combinations of $n$ variables. Each variable was selected from a pool of $m$ $3$-character strings, resulting in $m^n$ product combinations.
\textbf{Permutation} tasks required LLMs to generate all possible permutations of $m$ random strings, resulting in $m!$ permutations. Response coverage is quantified as the proportion of all possible enumerations returned by the model. Permutation results are presented in~\cref{tab:perm}, and combination results are included in the appendix.

These tasks represent enumerations with trivial predicates. Our PEA approach can be interpreted as synthesizing programs as
reasoning generators,
rather than embodying the reasoning itself. Program execution then yields concrete reasoning with precision and efficiency. Fine-grained complexity theory establishes hypothetical lower bounds for many computational problems, including polynomial-time solvable ones~\citep{Williams2019ONSF}, often resulting in more concise algorithmic descriptions compared to concrete reasons. For instance, a problem with a hypothetical lower bound of time complexity $O(n^3)$ maintains an algorithmic description (program representation) of $O(n)$, while concrete algorithmic reasoning takes $O(n^3)$ of time, which can be substantial for large problem instances, i.e.,
with large $n$.
Our LLM enumeration experiment exemplifies an extreme case where concrete evaluations grow exponentially, yet the programmatic representation remains highly succinct. Consequently, synthesizing the
reasoning generator
is significantly more efficient and simpler than 
performing
concrete reasoning in this scenario.

\begin{table}[th]
\begin{center}
\caption{Coverage of permutations: Percentage of total permutations returned by LLMs for sets of 4, 5, 6, and 7 random strings. Values represent means from three independent trials. For coverage below $10\%$, LLMs provide permutation-generation instructions rather than explicit results.}
\begin{tabular}{ ccccc } 
\toprule
\bf Model & \bf 4 & \bf 5 & \bf 6 & \bf 7 \\  
\hline
GPT-4o & 100\%  & 100\% & 84.9\% & 0.1\% \\  
\hline
o1-mini &  100\% &  100\% &  34.1\% &  0.2\% \\  
\hline
o1 &  100\% &  100\% &  13.6\% &  0.3\% \\  
  \bottomrule
\end{tabular}
\label{tab:perm}
\end{center}
\vspace{-0.3cm}
\end{table}

\subsection{SAT problem}\label{sec:sat}
The SAT problem involves determining a truth-value assignment that satisfies a given Boolean formula.

\paragraph{Formal Description} The SAT problem can be formally defined within the PEA framework as follows:
(i) The predicate is a Boolean formula
(for which a truth assignment evaluates to either True or False.) 
(ii) The enumeration comprises all possible truth assignments of the variables.
(iii) The aggregation determines whether any truth assignment in the enumeration satisfies the predicate.

\paragraph{Dataset}We use a Boolean-satisfiability dataset from~\citet{marino2024fastanalysisopenaio1preview}, comprising formulas in Conjunctive Normal Form (CNF). The dataset comprises 100 formulas each of 2-CNF, 3-CNF, and 4-CNF types, where clauses contain 2, 3, or 4 literals, respectively. While 2-CNF formula satisfiability is in class $\p$, 3-CNF and 4-CNF satisfiability problems are $\np$-complete. One 3-SAT formula is used for code semantic validation and excluded from the testing set, resulting in a total of 299 formulas for evaluation.

\paragraph{Results}Given the computational-complexity differences, results are presented separately: 2-CNF instances are aggregated in the appendix, while 3-CNF and 4-CNF instances are reported together in~\cref{tab:34-cnf}. This distinction reflects the $\p$-class nature of 2-CNF SAT problems and the $\np$-hardness of 3-CNF and 4-CNF SAT problems. PEA achieves perfect correctness in all cases, even when using the relatively weak reasoning model GPT-4o. Notably, all models perform exceptionally well on 2-CNF SAT problems, as detailed in the appendix.

\begin{table}
\begin{center}
\caption{Performance comparison of methods on 3-SAT and 4-SAT problems. Results are presented as $m+n$, where $m$ represents correctly solved satisfiable formulas (out of \textbf{172}) and $n$ denotes correctly identified unsatisfiable formulas (out of \textbf{27}).
}
\begin{tabular}{ ccccc } 
\toprule
\bf Model & \bf PEA & \bf CoT & \bf DQ  \\  
\hline
GPT-4o & 172+27  & 2+0 & 1+0  \\  
\hline
o1-mini &  172+27 &  46+11  &  63+13 \\  
\hline
o1 &  172+27 &  100+24 &  99+24  \\ 
  \bottomrule
\end{tabular}
\label{tab:34-cnf}
\end{center}
\vspace{-0.3cm}
\end{table}

\subsection{Game of 24}\label{sec:gon}
The Game of $24$ (\textbf{G24}) is a mathematical puzzle where participants are presented with four numbers and tasked with constructing an expression that evaluates to $24$. This expression can be formulated using the arithmetic operations of addition, subtraction, multiplication, and division, with the optional use of parentheses.

\paragraph{Formal description}The Game of $24$ can be formulated as follows: (i) The predicate evaluates whether an arithmetic expression composed of four given numbers equals $24$. (ii) The enumeration generates all valid expressions using $+$, $-$, $\times$, $\div$, and permissible parenthetical patterns. (iii) The aggregation determines if at least one expression evaluates to $n$ or if no such expression exists.

\paragraph{Dataset}We extract the most difficult $100$ instances of Game of $24$ numbers from the dataset of ToT~\citep{yao2023tree}. The difficulty of these instances is quantified by ToT's success rate. One example from the dataset, excluded from the $100$ selected instances, is used for code validation.

\begin{table}
\begin{center}
\caption{Performance comparison of methods on Game of 24. Results show the number of correctly assembled expressions out of \textbf{100} valid instances. ToT with o1 and o1-mini were omitted due to excessive query time (around 1 hour per instance) and poor performance (0 out of 10 in preliminary testing).}
\begin{tabular}{ccccc} 
\toprule
\bf Model & \bf PEA & \bf CoT & \bf DQ & \bf ToT \\  
\hline
GPT-4o & 100  & 2 & 2 & 4\\  
\hline
o1-mini & 100  &  29  & 44  & -- \\  
\hline
o1 & 100 &  60 &  50  & -- \\ 
  \bottomrule
\end{tabular}
\label{tab:g24}
\end{center}
\vspace{-0.3cm}
\end{table}

\paragraph{Results}The results of the Game of $24$ are presented in~\cref{tab:g24}. PEA demonstrates perfect performance across all models, while the most advanced o1 model, even when equipped with CoT, resolves only $60$ instances. 

\subsection{Planning problems}\label{sec:planning}
We use a recent planning benchmark dataset from~\citet{valmeekam2023planbench}, designed to assess LLMs' reasoning capabilities in solving planning problems. The benchmark comprises several variants of two primary tasks: Blocksworld (\textbf{BW}) and Logistics (\textbf{LOGI}).

The Blocksworld task is a planning problem where participants are presented with several colored blocks on a table. The task requires executing valid actions to move these blocks. Each problem instance provides an initial block configuration and goal conditions representing desired block-stacking conditions. Participants must devise a feasible plan, comprising a sequence of actions, to achieve the goal state. Our evaluation uses a more challenging variant, the optimal-cost Blocksworld problem, which necessitates finding a minimum-cost action sequence.

The Logistics task involves a planning problem featuring packages, vehicles, and cities with various locations. Vehicles, including trucks and airplanes, must adhere to specific rules for movement and package handling (loading/unloading). Each problem instance describes initial settings and package destinations. Participants are required to formulate a viable plan for vehicle actions to meet the specified goals. Due to inconsistencies in the cost-optimal logistics dataset, this study evaluates performance based on the satisfiable version of the problem, focusing on identifying any valid plan rather than the optimal solution.


\paragraph{Formal description} We now formalize the planning problems in the PEA framework.
For cost-optimal \textbf{BW}: (i) The predicate is to determine if a given initial block configuration and action list lead to the goal conditions. (ii) The enumeration sequentially generates all valid action lists from the current state. (iii) The aggregation evaluates if the action list, ordered by ascending cost, satisfies the predicate, which effectively implements a breadth-first search.
For valid \textbf{LOGI}: (i) The predicate is to determine if a given initial package configuration with a list of vehicle actions can reach the goal state. (ii) The enumeration sequentially generates all valid vehicle-action lists from the current vehicle and package state. (iii) The aggregation evaluates if the action list satisfies the predicate.

\paragraph{Dataset} For the cost-optimal Blocksworld task, we use all $500$ task instances from the Plan-Optimality dataset. Each instance involves blocks of various colors, with minimal step requirements ranging from a few to a dozen. The instances include a natural-language description of rules, a one-shot example (comprising initial state, goal conditions, and plan), and the task question.

For the valid Logistics task, we employ all $285$ task instances from the Plan-Generation dataset. Each instance involves a specific configuration of cities, locations, vehicles, and packages, with plan actions ranging from a few to approximately forty. Similar to BW, each instance contains a natural-language description of rules, a one-shot example, and the task question.

The dataset provides ground truth for each task instance and a reference validator for plan verification.

\begin{table}
\begin{center}
    \caption{Performance comparison of methods on cost-optimal Blocksworld task: Table displays the number of correctly solved instances out of \textbf{500} total tasks for various methods.}
    \begin{tabular}{ ccccc } 
        \toprule \bf Model & \bf PEA & \bf CoT & \bf DQ  \\
        \hline GPT-4o & 500 & 115 & 94  \\  
        \hline o1-mini & 500 & 185 &  182 \\  
        \hline o1 & 500 & 452 &  324  \\ 
        \bottomrule
    \end{tabular}
    \label{tab:bw}
\end{center}
\vspace{-0.3cm}
\end{table}

\paragraph{Results} The results for the cost-optimal Blocksworld task are presented in~\cref{tab:bw}. Notably, the o1 model demonstrates superior performance on this task compared to other models. However, when using PEA, all models achieve perfect task completion.

\Cref{tab:logi} summarizes the results for the valid Logistics task. The optimization option is activated for PEA due to the extensive search space. While GPT-4o generates only unoptimized code, resulting in frequent $30$-second timeouts, it still successfully solves $78$ instances, outperforming its direct-query baselines. In contrast, both o1 and o1-mini models demonstrate superior performance, successfully synthesizing optimized programs through PEA and solving all Logistics instances.

\begin{table}
\begin{center}
    \caption{Performance comparison on valid Logistics task: Results present the number of correctly solved instances out of \textbf{285} tasks. Note: GPT-4o PEA generates unoptimized code; $207$ instances reach $30$-second timeout and are terminated.}
    \begin{tabular}{ ccccc } 
        \toprule \bf Model & \bf PEA & \bf CoT & \bf DQ  \\
        \hline GPT-4o & 78 & 44 & 28  \\  
        \hline o1-mini & 285 & 60 &  65 \\  
        \hline o1 & 285 & 205 &  163  \\ 
        \bottomrule
    \end{tabular}
    \label{tab:logi}
\end{center}
\vspace{-0.3cm}
\end{table}

\subsection{Efficiency}\label{sec:eff}
To evaluate the efficiency of each prompt method, we measure the per-instance solving time. For  PEA, the initial program-synthesis cost is amortized across all instances, because it occurs only at the problem's outset. \Cref{tab:efficiency} presents the per-instance computation costs. Notably, the PEA method not only enhances problem-solving accuracy but also demonstrates reduced computation time in most cases across all models tested.
\begin{table}
\begin{center}
    \caption{Per-instance computation time (in seconds) for each problem-solving approach:
While o1 achieves high accuracy, the PEA approach demonstrates superior accuracy with significantly lower computational cost due to program reuse across multiple instances.}
    \label{tab:efficiency}
    \begin{tabular}{ ccccc } 
        \toprule \bf Method(Model) & \bf SAT & \bf G24 & \bf BW  & \bf LOGI  \\
        \hline PEA (GPT-4o) & 0.03 & 0.1 & 2.3 & 23.3 \\  
        \hline COT (GPT-4o) & 7.3 & 7.1 & 8.2 & 13.6 \\
        
        \hline DQ (GPT-4o) &  3.5 & 4.5 & 3.6 & 8.0 \\
        \midrule
        \hline PEA (o1-mini) & 0.1 & 0.2 & 2.8 & 7.9 \\  
        \hline COT (o1-mini) & 16.6 & 11.4 & 18.8 & 55.0 \\
        \hline DQ (o1-mini) & 23.6 & 16.9 & 15.3 & 27.8 \\
        \midrule
        \hline PEA (o1) & 0.1 & 0.9 & 2.6 & 15.8 \\  
        \hline COT (o1) & 206.2 & 56.8 & 98.7 & 181.5 \\
        \hline DQ (o1) & 159.7 & 100.8 & 134.5 & 96.8 \\
        \bottomrule
    \end{tabular}
\end{center}
\vspace{-0.3cm}
\end{table}

\section{Conclusion}This paper introduces the PEA framework, which formally characterizes computational reasoning tasks and enhances LLMs' reasoning capabilities through their coding proficiency. Our central premise posits that many computational problems have inherent lower bounds, and synthesizing programs as reasoning generators is often more efficient and simpler than performing concrete reasoning.





\section*{Acknowledgment}
\thanks{Supported, in part,
by a gift from Rajiv and Ritu Batra; and by NSF under grants CCF-2212558, CCF-FMiTF-1836978, DARPA under agreement number 885000 and ONR N00014-21-1-2492.
%
Any opinions, findings, and conclusions or recommendations expressed in this publication are those of the authors, and do not necessarily reflect the views of the sponsoring entities.
}

\bibliographystyle{plainnat}
\bibliography{paper}
    
\appendix

\newpage
\appendix

\lstdefinestyle{mystyle}{
  stringstyle=\color{codepurple},
  basicstyle=\ttfamily\footnotesize,
  escapeinside={@}{@},
  breakatwhitespace=false,         
  breaklines=true,                 
  captionpos=b,                    
  keepspaces=true,                 
  showspaces=false,                
  showstringspaces=false,
  showtabs=false,                  
  tabsize=2
}

\lstset{style=mystyle}

\onecolumn
This appendix presents supplementary materials. \Cref{sec:pal} provides a comparison between PEA and PAL. \Cref{sec:eval} contains additional empirical results omitted from the main text. \Cref{sec:prompts} details the prompts utilized for PEA in our evaluation.

\section{PEA vs PAL}\label{sec:pal}
PAL~\citep{pal} and our PEA framework framework both utilize code generation to enhance LLM performance on reasoning tasks. From a first-order logic perspective, PAL addresses less complex tasks involving direct concrete predicate evaluation without quantifiers, while PEA is designed to handle more complex logical structures. To illustrate, consider the following example of a PAL task:
\begin{tcolorbox}[title=PAL Example]
Q: Olivia has $23$. She bought five bagels for $3$ each. How much money does she have left?

\# solution in Python:
\begin{lstlisting}[language=Python]

def solution():
    """Olivia has $23. She bought five bagels for $3 each. 
    How much money does she have left?"""
    money_initial = 23
    bagels = 5
    bagel_cost = 3
    money_spent = bagels * bagel_cost
    money_left = money_initial - money_spent
    result = money_left
    return result
\end{lstlisting}

\end{tcolorbox}
In such problems, the solution merely requires extracting predicates from the natural language description and evaluating them directly. This contrasts with the quantified predicate problems primarily addressed by PEA, which necessitate more sophisticated enumeration strategies.

\section{Additional empirical results}\label{sec:eval}
\subsection{Cartesian product combination}
The results of the Cartesian product-combination evaluation are presented in~\cref{tab:prod}. These findings demonstrate a similar pattern to those observed in the permutation evaluation.

\begin{table}[!ht]
\begin{center}
\caption{Coverage of Cartesian product combinations: Percentage of total product combinations returned in LLM responses. Notation $(m, n)$ represents $n$ variables, each selecting from a pool of $m$ random string values, yielding $m^n$ product combinations. Values reported are means from three independent trials. For coverage below $10\%$, LLMs provide Cartesian product generation instructions rather than explicit results.
}
\begin{tabular}{ ccccc } 
\toprule
\bf Model & \bf (4,4) & \bf (5,4) & \bf (4,5) & \bf (5,5) \\  
\hline
GPT-4o & 100\%  & 100\% & 57.8\% & 26.8\% \\  
\hline
o1-mini &  100\% &  32.5\% &  1.3\% &  0.4\% \\  
\hline
o1 &  100\% &  67.7\% &  6.1\% &  0.9\% \\ 
  \bottomrule
\end{tabular}
\label{tab:prod}
\end{center}
\vspace{-0.3cm}
\end{table}

\subsection{2-SAT}
The results for the 2-SAT problem are presented in~\cref{tab:2-cnf}. Given that 2-SAT is a $\p$-class problem with efficient algorithmic solutions, it is noteworthy that all models demonstrated commendable performance in this evaluation.

\begin{table}[h!]
\begin{center}
\caption{Performance comparison of methods on 2-SAT problems. Results show the number of correctly solved formulas out of \textbf{100} satisfiable 2-CNF instances.
}
\begin{tabular}{ ccccc } 
\toprule
\bf Model & \bf PEA & \bf CoT & \bf DQ  \\  
\hline
GPT-4o & 100  & 88 & 77  \\  
\hline
o1-mini & 100 &  100  & 100  \\  
\hline
o1 &  100 &  100 &  100  \\ 
  \bottomrule
\end{tabular}
\label{tab:2-cnf}
\end{center}
\vspace{-0.3cm}
\end{table}

\subsection{Deepseek preliminary results}

Preliminary small-scale experiments were conducted utilizing DeepSeek models, specifically DeepSeek-V3 (\textit{deepseek-chat}) and DeepSeek-R1 (\textit{deepseek-reasoner}), for Blocksworlds and Logistics planning tasks via the DeepSeek API~\citep{deepseekai2024deepseekv3technicalreport,deepseekai2025deepseekr1incentivizingreasoningcapability}. PEA when applied to these models, necessitated multiple iterations to synthesize programs meeting sanity check criteria. However, the generated programs failed to achieve sufficient accuracy. While the web interface demonstrated rapid response times for direct queries and CoT approaches, API requests exhibited slow performance and poor connection stability. The obtained results did not match the quality of those from OpenAI o1. Consequently, despite intentions to conduct a comprehensive evaluation, the decision was made to defer full-scale testing pending improved service reliability.

\newcommand{\hlc}[2][yellow]{{%
    \colorlet{foo}{#1}%
    \sethlcolor{foo}\hl{#2}}%
}
\section{PEA Prompts}\label{sec:prompts}
This section presents the core PEA prompts utilized in our evaluation, \hlc[pink]{highlighting in pink} the PEA components. For prompts incorporating enumeration optimization strategies, the strategies are \hlc[yellow]{highlighted in yellow} and noted to be LLM-generated rather than user-crafted. The majority of the prompt content for planning problems is directly from the problem descriptions provided in the dataset.
\begin{tcolorbox}[title=SAT Prompt]
\begin{lstlisting}
You are given a boolean formula with 10 boolean variables and conjunctive normal form. Find a satisfiable assignment for this formula and return the assignment to me as a list of 10 True/False values in a bracket [], for example, [True, False, True, True, False, False, False, True, True, False]; If this formula is not satisfiable, then return me UNSAT.

Now the tasks are to generate three Python functions to answer this question. We will use a Predicate-Enumeration-Aggregation template to solve this problem.

@\hlc[pink]{For the predicate, implement a function "evaluate\_formula" to with inputs: a python function "formula" and a bool variable list "vals". Evaluate "formula" with "vals" and return True or False.}@

@\hlc[pink]{For the enumeration, implement a function "enumerate\_boolean" that generates all possible enumerations of 10 boolean variables.}@

@\hlc[pink]{For the aggregation, implement a function "can\_evaluate" to find either valid truth assignment or UNSAT. "can\_evaluate" takes a python function "formula" as input and use the previously generated functions by checking whether any of the boolean assignment from "enumerate\_boolean" can satisfy formula by passing the assignment and "formula" to "evaluate\_formula". If such an assignment exists, print out the assignment and return True; else, return False.}@

You may add comments to the generated code to show how the implementation reflects the instruction.
\end{lstlisting}
\end{tcolorbox}

\begin{tcolorbox}[title=G24 Prompt]
\begin{lstlisting}
Given any four positive numbers which may have repetitions; Using just the +, -, *, and / operators; and the possible use of parenthesis, (), show how to make an answer of 24. You will use each number once and only once. Return me the arithmetic expression in a bracket, for example,[6+2*(4+5)]; if it cannot make an answer 24, then tell me it cannot.

Now the tasks are to generate Python functions to answer this question. We will use a Predicate-Enumeration-Aggregation template to solve this problem.

@\hlc[pink]{For the predicate, implement a function "evaluate\_to\_24" to with input: an expression "expression" and check whether the expression can be evaluated to 24. Notice arithemtical errors like divide-by-zero and evaluation precision. Return True if it is evaluated to 24; or False otherwise.}@

@\hlc[pink]{For the enumeration, implement a function "generate\_expressions" that with input: four numbers "n1, n2, n3, n4" and generates all possible arithmetical expressions of the four input numbers, that uses +, -, *, and /; and all possible parenthesis patterns.}@

@\hlc[pink]{For the aggregation, implement a function "can\_evaluate" to find either a valid 24 expression or impossible. "can\_evaluate" takes four numbers as input and uses the previously generated functions by checking whether any of the expressions from "generate\_expressions" can be evaluated to 24 by calling "evaluate\_to\_24".
If yes, print the expression and return True;
If no expression evaluates to 24, then return False.}@

You may add comments to the generated code to show how the implementation reflects the instruction.
\end{lstlisting}

\end{tcolorbox}

\begin{tcolorbox}[breakable, title=Blocksworld Prompt]
\begin{lstlisting}
I am playing with a set of blocks where I need to arrange the blocks into stacks. Here are the actions I can do:

Pick up a block. It takes 1 minute to pick up a block.
Unstack a block from on top of another block. It takes 1 minute to unstack a block from on top of another block.
Put down a block. It takes 1 minute to put down a block.
Stack a block on top of another block. It takes 1 minute to stack a block on top of another block.

I have the following restrictions on my actions:
I can only pick up or unstack one block at a time.
I can only pick up or unstack a block if my hand is empty.
I can only pick up a block if the block is on the table and the block is clear. A block is clear if the block has no other blocks on top of it and if the block is not picked up.
I can only unstack a block from on top of another block if the block I am unstacking was really on top of the other block.
I can only unstack a block from on top of another block if the block I am unstacking is clear.
Once I pick up or unstack a block, I am holding the block.
I can only put down a block that I am holding.
I can only stack a block on top of another block if I am holding the block being stacked.
I can only stack a block on top of another block if the block onto which I am stacking the block is clear.
Once I put down or stack a block, my hand becomes empty.
Once you stack a block on top of a second block, the second block is no longer clear.

[STATEMENT]
As initial conditions I have that, the red block is clear, the blue block is clear, the yellow block is clear, the hand is empty, the blue block is on top of the orange block, the red block is on the table, the orange block is on the table and the yellow block is on the table.
My goal is to have that the orange block is on top of the blue block. I want to minimize the time taken to achieve my goal.

My plan is as follows:

[PLAN]
unstack the blue block from on top of the orange block
put down the blue block
pick up the orange block
stack the orange block on top of the blue block
[PLAN END]
The total time to execute the plan is 4 minutes.

[STATEMENT]
As initial conditions I have that, the red block is clear, the yellow block is clear, the hand is empty, the red block is on top of the blue block, the yellow block is on top of the orange block, the blue block is on the table and the orange block is on the table.
My goal is to have that the orange block is on top of the red block. I want to minimize the time taken to achieve my goal.
My plan is as follows:

[PLAN]

@\hlc[pink]{Generate a class `BlockWorldState` to represent the state of the blocks world. The state includes the positions of the blocks, the clear state of the blocks, and the state of the hand.}@

@\hlc[pink]{For the predicate, check function `check\_goal` to determine whether a state satified the goal conditions.}@

@\hlc[pink]{For the enumeration, implement function `generate\_actions` to generate valid actions at the current state. You can almost have four possible actions but should filter out with given restrictions. Also implement function `apply\_action` to apply an action to the current state. The function should return a new state after applying the action.}@

@\hlc[pink]{For the aggregation, implement a function `find\_plan` to find a plan to reach the target goal. The function should take the initial state and the goal  conditions as input and return a list of actions. For the implementation, the function should use BFS. The function should use functions generated above. Also implement a function `verify\_plan` to verify a given plan. The function should take the initial state, the goal conditions, and the plan as input and return a boolean value.
}@

\end{lstlisting}
\end{tcolorbox}

\begin{tcolorbox}[breakable, title=Logistics Prompt]
\begin{lstlisting}
I have to plan logistics to transport packages within cities via trucks and between cities via airplanes. Locations within a city are directly connected (trucks can move between any two such locations), and so are the cities. In each city there is exactly one truck and each city has one location that serves as an airport.
Here are the actions that can be performed:

Load a package into a truck. For example, load package_1 into truck_1 at location_1_1.
Load a package into an airplane. For example, load package_1 into airplane_1 at location_1_1.
Unload a package from a truck. For example, unload package_1 from truck_1 at location_1_1.
Unload a package from an airplane. For example, unload package_1 from airplane_1 at location_1_1.
Drive a truck from one location to another location. For example, drive truck_1 from location_1_1 to location_1_2 in city_1.
Fly an airplane from one city to another city. For example, fly airplane_1 from location_1_1 to location_2_1. Here location_1_1 is the airport in city_1 and location_2_1 is the airport in city_2.

The following are the restrictions on the actions:
A package can be loaded into a truck only if the package and the truck are in the same location.
Once a package is loaded into a truck, the package is not at the location and is in the truck.   
A package can be loaded into an airplane only if the package and the airplane are in the same location.
Once a package is loaded into an airplane, the package is not at the location and is in the airplane.
A package can be unloaded from a truck only if the package is in the truck.
Once a package is unloaded from a truck, the package is not in the truck and is at the location of the truck.
A package can be unloaded from an airplane only if the package in the airplane.
Once a package is unloaded from an airplane, the package is not in the airplane and is at the location of the airplane.   
A truck can be driven from one location to another if the truck is at the from-location and both from-location and to-location are locations in the same city.
Once a truck is driven from one location to another, it is not at the from-location and is at the to-location.
An airplane can be flown from one city to another if the from-location and the to-location are airports and the airplane is at the from-location.
Once an airplane is flown from one city to another the airplane is not at the from-location and is at the to-location.

[STATEMENT]
As initial conditions I have that, location_0_0 is an airport, location_1_0 is an airport, airplane_0 is at location_1_0, package_0 is at location_1_0, truck_0 is at location_0_0, truck_1 is at location_1_0, location_0_0 is in the city city_0, location_0_1 is in the city city_0, location_1_0 is in the city city_1 and location_1_1 is in the city city_1.
My goal is to have that package_0 is at location_1_1.

My plan is as follows:

[PLAN]
load package_0 into truck_1 at location_1_0
drive truck_1 from location_1_0 to location_1_1 in city_1
unload package_0 from truck_1 at location_1_1
[PLAN END]

[STATEMENT]
As initial conditions I have that, location_0_0 is an airport, location_1_0 is an airport, airplane_0 is at location_0_0, package_0 is at location_0_1, truck_0 is at location_0_0, truck_1 is at location_1_1, location_0_0 is in the city city_0, location_0_1 is in the city city_0, location_1_0 is in the city city_1 and location_1_1 is in the city city_1.
My goal is to have that package_0 is at location_1_0.

My plan is as follows:

[PLAN]

@\hlc[pink]{Generate a class `LogisticsState` to represent the state of the world. Remember to import the deepcopy if you use it.}@

@\hlc[pink]{For the predicate, check function `check\_goal` to determine whether a state satified the goal conditions.}@

@\hlc[pink]{For the enumeration, implement the function `generate\_actions` to generate valid actions at the current state.}@

@\hlc[pink]{For the aggregation, implement a function `find\_plan` to find a plan to reach the target goal. The function should take the initial state and the goal conditions as input and return a list of actions. For the implementation, the function should use BFS. The function should use functions generated above. Also implement a function `verify\_plan` to verify a given plan. The function should take the initial state, the goal conditions, and the plan condition as input and return a boolean value.}@
\end{lstlisting}
\end{tcolorbox}

\begin{tcolorbox}[breakable, title=Logistics Prompt for Optimization Strategies]
\begin{lstlisting}
I have to plan logistics to transport packages within cities via trucks and between cities via airplanes. Locations within a city are directly connected (trucks can move between any two such locations), and so are the cities. In each city there is exactly one truck and each city has one location that serves as an airport.
Here are the actions that can be performed:

Load a package into a truck. For example, load package_1 into truck_1 at location_1_1.
Load a package into an airplane. For example, load package_1 into airplane_1 at location_1_1.
Unload a package from a truck. For example, unload package_1 from truck_1 at location_1_1.
Unload a package from an airplane. For example, unload package_1 from airplane_1 at location_1_1.
Drive a truck from one location to another location. For example, drive truck_1 from location_1_1 to location_1_2 in city_1.
Fly an airplane from one city to another city. For example, fly airplane_1 from location_1_1 to location_2_1. Here location_1_1 is the airport in city_1 and location_2_1 is the airport in city_2.

The following are the restrictions on the actions:
A package can be loaded into a truck only if the package and the truck are in the same location.
Once a package is loaded into a truck, the package is not at the location and is in the truck.   
A package can be loaded into an airplane only if the package and the airplane are in the same location.
Once a package is loaded into an airplane, the package is not at the location and is in the airplane.
A package can be unloaded from a truck only if the package is in the truck.
Once a package is unloaded from a truck, the package is not in the truck and is at the location of the truck.
A package can be unloaded from an airplane only if the package in the airplane.
Once a package is unloaded from an airplane, the package is not in the airplane and is at the location of the airplane.   
A truck can be driven from one location to another if the truck is at the from-location and both from-location and to-location are locations in the same city.
Once a truck is driven from one location to another, it is not at the from-location and is at the to-location.
An airplane can be flown from one city to another if the from-location and the to-location are airports and the airplane is at the from-location.
Once an airplane is flown from one city to another the airplane is not at the from-location and is at the to-location.

[STATEMENT]
As initial conditions I have that, location_0_0 is an airport, location_1_0 is an airport, airplane_0 is at location_1_0, package_0 is at location_1_0, truck_0 is at location_0_0, truck_1 is at location_1_0, location_0_0 is in the city city_0, location_0_1 is in the city city_0, location_1_0 is in the city city_1 and location_1_1 is in the city city_1.
My goal is to have that package_0 is at location_1_1.

My plan is as follows:

[PLAN]
load package_0 into truck_1 at location_1_0
drive truck_1 from location_1_0 to location_1_1 in city_1
unload package_0 from truck_1 at location_1_1
[PLAN END]

[STATEMENT]
As initial conditions I have that, location_0_0 is an airport, location_1_0 is an airport, airplane_0 is at location_0_0, package_0 is at location_0_1, truck_0 is at location_0_0, truck_1 is at location_1_1, location_0_0 is in the city city_0, location_0_1 is in the city city_0, location_1_0 is in the city city_1 and location_1_1 is in the city city_1.
My goal is to have that package_0 is at location_1_0.

My plan is as follows:

[PLAN]

@\hlc[pink]{Generate a class `LogisticsState` to represent the state of the world. Remember to import the deepcopy if you use it.}@

@\hlc[pink]{For the predicate, check function `check\_goal` to determine whether a state satified the goal conditions.}@

@\hlc[pink]{For the enumeration, implement the function `generate\_actions` to generate valid actions at the current state.}@

@\hlc[pink]{For the aggregation, implement a function `find\_plan` to find a plan to reach the target goal. The function should take the initial state and the goal conditions as input and return a list of actions. For the implementation, the function should use BFS. The function should use functions generated above. Also implement a function `verify\_plan` to verify a given plan. The function should take the initial state, the goal conditions, and the plan condition as input and return a boolean value.}@

@\hlc[pink]{Instead of generating the above code, for function generate\_actions,  generate a structured and succinct instruction description for generating one deterministic action for reaching packages' goals, trying to think about all possible actions according to the package, truck and airplane's location and remove unhelpful actions.  I will insert it in the later prompt for code generation so you don't need to have a summary.}@

\end{lstlisting}
\end{tcolorbox}

\begin{tcolorbox}[breakable, title=Logistics Prompt with Enumeration Optimization Strategies]
\begin{lstlisting}
I have to plan logistics to transport packages within cities via trucks and between cities via airplanes. Locations within a city are directly connected (trucks can move between any two such locations), and so are the cities. In each city there is exactly one truck and each city has one location that serves as an airport.
Here are the actions that can be performed:

Load a package into a truck. For example, load package_1 into truck_1 at location_1_1.
Load a package into an airplane. For example, load package_1 into airplane_1 at location_1_1.
Unload a package from a truck. For example, unload package_1 from truck_1 at location_1_1.
Unload a package from an airplane. For example, unload package_1 from airplane_1 at location_1_1.
Drive a truck from one location to another location. For example, drive truck_1 from location_1_1 to location_1_2 in city_1.
Fly an airplane from one city to another city. For example, fly airplane_1 from location_1_1 to location_2_1. Here location_1_1 is the airport in city_1 and location_2_1 is the airport in city_2.

The following are the restrictions on the actions:
A package can be loaded into a truck only if the package and the truck are in the same location.
Once a package is loaded into a truck, the package is not at the location and is in the truck.   
A package can be loaded into an airplane only if the package and the airplane are in the same location.
Once a package is loaded into an airplane, the package is not at the location and is in the airplane.
A package can be unloaded from a truck only if the package is in the truck.
Once a package is unloaded from a truck, the package is not in the truck and is at the location of the truck.
A package can be unloaded from an airplane only if the package in the airplane.
Once a package is unloaded from an airplane, the package is not in the airplane and is at the location of the airplane.   
A truck can be driven from one location to another if the truck is at the from-location and both from-location and to-location are locations in the same city.
Once a truck is driven from one location to another, it is not at the from-location and is at the to-location.
An airplane can be flown from one city to another if the from-location and the to-location are airports and the airplane is at the from-location.
Once an airplane is flown from one city to another the airplane is not at the from-location and is at the to-location.

[STATEMENT]
As initial conditions I have that, location_0_0 is an airport, location_1_0 is an airport, airplane_0 is at location_1_0, package_0 is at location_1_0, truck_0 is at location_0_0, truck_1 is at location_1_0, location_0_0 is in the city city_0, location_0_1 is in the city city_0, location_1_0 is in the city city_1 and location_1_1 is in the city city_1.
My goal is to have that package_0 is at location_1_1.

My plan is as follows:

[PLAN]
load package_0 into truck_1 at location_1_0
drive truck_1 from location_1_0 to location_1_1 in city_1
unload package_0 from truck_1 at location_1_1
[PLAN END]

[STATEMENT]
As initial conditions I have that, location_0_0 is an airport, location_1_0 is an airport, airplane_0 is at location_0_0, package_0 is at location_0_1, truck_0 is at location_0_0, truck_1 is at location_1_1, location_0_0 is in the city city_0, location_0_1 is in the city city_0, location_1_0 is in the city city_1 and location_1_1 is in the city city_1.
My goal is to have that package_0 is at location_1_0.

My plan is as follows:

[PLAN]

@\hlc[pink]{Generate a class `LogisticsState` to represent the state of the world. Remember to import the deepcopy if you use it.}@

@\hlc[pink]{For the predicate, check function `check\_goal` to determine whether a state satified the goal conditions.}@

@\hlc[yellow]{For the enumeration, implement the function `generate\_actions` to generate useful and valid actions at the current state. The function should follow these rules:}@

@\hlc[yellow]{
Below is a concise step‐by‐step procedure for selecting **one** “helpful” action that moves at least one package closer to its goal location (no unnecessary detours). You can integrate this logic into `generate\_actions` to produce exactly one deterministic next action (rather than enumerating everything) that progresses a single package:}@

@\hlc[yellow]{
\#\#\# 1. Identify a Package to Move
}@
@\hlc[yellow]{1. Gather all packages **not** at their goal locations.}@
@\hlc[yellow]{2. Pick **one** such package (e.g., the first in a list).}@

@\hlc[yellow]{
\#\#\# 2. Check If Package Is in the Correct City}@

@\hlc[yellow]{1. **If the package is not in its goal city**:}@

    @\hlc[yellow]{1. **Not Loaded,** at some location (possibly an airport):}@
     @\hlc[yellow]{- If a **truck** is co‐located and this location is **not** an airport, use \`{}['load-truck', package, truck, location]\`{}.}@

     @\hlc[yellow]{- Else if an **airplane** is co‐located and this location **is** an airport, use \`{}['load-airplane', package, airplane, airport\_location]\`{}.}@

     @\hlc[yellow]{- If neither vehicle is co‐located, pick a valid action to bring the vehicle to the package (e.g., drive the city’s truck over, or fly the airplane in if it is an airport) — returning the **first** such valid movement action.}@

   @\hlc[yellow]{2. **Loaded in a truck**, but not at the airport:}@
     @\hlc[yellow]{- Drive that truck toward its city’s airport to prepare for intercity shipment:  }@
        @\hlc[yellow]{\`{}['drive-truck', truck, current\_loc, airport\_loc, city]\`{}.}@
        
   @\hlc[yellow]{3. **Loaded in an airplane**:}@
     @\hlc[yellow]{- Fly to the destination city’s airport:  }@
        @\hlc[yellow]{\`{}['fly-airplane', airplane, current\_airport, destination\_airport]\`{}.}@

@\hlc[yellow]{2. **If the package is in its correct city, but not at its goal location**:}@
   @\hlc[yellow]{1. **Not Loaded** and co‐located with the city’s truck:}@
     @\hlc[yellow]{- Load it:  
        \`{}['load-truck', package, truck, location]\`{}.}@
     @\hlc[yellow]{- If the truck is **not** at the same location yet, drive the truck there first.}@

   @\hlc[yellow]{2. **Loaded in the city’s truck**:}@
     @\hlc[yellow]{- Drive the truck to the package’s goal location: \`{}['drive-truck', truck, current\_loc, goal\_loc, city]\`{}.}@

   @\hlc[yellow]{3. **In the truck at goal location**:}@
     @\hlc[yellow]{ Unload it:  
        \`{}['unload-truck', package, truck, goal\_loc]\`{}.}@

@\hlc[yellow]{
\#\#\# 3. Return the Single Action
- As soon as you identify a valid step (load, drive, fly, or unload) that progresses your chosen package **toward** its goal, **return** that action immediately (rather than continuing to look for others). This ensures only the “helpful” action is generated.}@

   @\hlc[yellow]{- As soon as you identify a valid step (load, drive, fly, or unload) that progresses your chosen package **toward** its goal, **return** that action immediately (rather than continuing to look for others). This ensures only the “helpful” action is generated.}@

@\hlc[yellow]{This procedure avoids superfluous moves: it always moves a relevant vehicle/piece of cargo step by step closer to the correct final location, ignoring any action that does not help a package reach its goal.}@

@\hlc[pink]{For the aggregation, implement a function `find\_plan` to find a plan to reach the target goal. The function should take the initial state and the goal conditions as input and return a list of actions. For the implementation, the function should use BFS. The function should use functions generated above. Also implement a function `verify\_plan` to verify a given plan. The function should take the initial state, the goal conditions, and the plan condition as input and return a boolean value. }@

Let's think step by step.
\end{lstlisting}
\end{tcolorbox}

\end{document}